\documentclass[letterpaper, 10 pt, conference]{ieeeconf}  

\IEEEoverridecommandlockouts                              

\usepackage{graphicx} 
\usepackage{subfigure}
\usepackage{amsmath}
\usepackage{amssymb}
\usepackage{xcolor}

\usepackage{tikz}
\usepackage{textcomp}
\usepackage{hyperref}
\usepackage{lipsum}

\newcommand\copyrighttext{%
  \footnotesize \textcopyright \ 2019 IEEE. Personal use of this material is permitted.
  Permission from IEEE must be obtained for all other uses, in any current or future
  media, including reprinting/republishing this material for advertising or promotional
  purposes, creating new collective works, for resale or redistribution to servers or
  lists, or reuse of any copyrighted component of this work in other works.}
\newcommand\copyrightnotice{%
\begin{tikzpicture}[remember picture,overlay]
\node[anchor=south,yshift=10pt] at (current page.south) {\fbox{\parbox{\dimexpr\textwidth-\fboxsep-\fboxrule\relax}{\copyrighttext}}};
\end{tikzpicture}%
}

\title{\LARGE \bf
Learning to Dynamically Coordinate Multi-Robot Teams in Graph Attention Networks
}

\author{Zheyuan Wang$^{1}$ and Matthew Gombolay$^{1}$
\thanks{*This work was supported by the Office of Naval Research under grant GR10006659 and Lockheed Martin Corporation under grant GR00000509.}
\thanks{$^{1}$Zheyuan Wang and Matthew Gombolay are affiliated with the Georgia Institute of Technology, Atlanta, GA 30332, USA
        {\tt\small \{pjohnwang,mgombolay3\}@gatech.edu}}%
}

\begin{document}

\maketitle
\copyrightnotice
\thispagestyle{empty}
\pagestyle{empty}

\begin{abstract}

Increasing interest in integrating advanced robotics within manufacturing has spurred a renewed concentration in developing real-time scheduling solutions to coordinate human-robot collaboration in this environment. Traditionally, the problem of scheduling agents to complete tasks with temporal and spatial constraints has been approached either with exact algorithms, which are computationally intractable for large-scale, dynamic coordination, or approximate methods that require domain experts to craft heuristics for each application. We seek to overcome the limitations of these conventional methods by developing a novel graph attention network formulation to automatically learn features of scheduling problems to allow their deployment. To learn effective policies for combinatorial optimization problems via machine learning, we combine imitation learning on smaller problems with deep Q-learning on larger problems, in a non-parametric framework, to allow for fast, near-optimal scheduling of robot teams. We show that our network-based policy finds at least twice as many solutions over prior state-of-the-art methods in all testing scenarios.
\end{abstract}

\section{INTRODUCTION}

Advances in robotic technology are enabling the introduction of mobile robots into manufacturing environments alongside human workers. By removing the cage around traditional robot platforms and integrating dynamic, final assembly operations with human-robot teams, manufacturers can see improvements in reducing a factory's footprint and environmental costs, as well as increased productivity~\cite{heyer2010human}. For human workspaces associated with final assembly, tasks need to be quickly allocated and sequenced (i.e., scheduled) among a set of robotic agents to achieve a high-quality schedule with respect to the application-specific objective function while satisfying the temporal constraints (i.e., upper and lower bound deadline, wait, and task duration constraints), as well as spatial constraints on agent proximity for safe and efficient collaboration with human workers. The problem of resource optimization is made difficult by the inter-coupled constraints requiring a joint schedule rather than allowing each agent to compute their work plans independently. 

Conventional approaches to scheduling typically involve formulating the problem as a mathematical program and leveraging commercial solvers or developing custom-made approximate and meta-heuristic techniques. Exact algorithms aim to find the optimal schedule based on enumeration or branch-and-bound, making them computationally expensive and unable to scale to large, real-time scheduling. Exact methods often rely on hand-crafted, ``warm-start'' heuristics unique to each application. Alternatively, heuristic approaches are lightweight and often effective; however, designing application-specific heuristics requires extracting and encoding domain-expert knowledge through interviews and trial-and-error-based research, a process which leaves much to be desired. Furthermore, accurately and efficiently extracting this knowledge remains an open problem~\cite{raghavan2006active}.

To overcome the limitations of prior work, we build on promising developments in deep-learning-based architectures (i.e., graph neural networks) to learn heuristics for combinatorial problems. Analogous to the convolutional neural networks for feature-learning in images, graph neural networks are able to hierarchically learn high-level representations of graph structures through convolutions and backpropagation. Yet, these approaches have only been developed for simpler scheduling problems, such as the traveling salesman problem~\cite{khalil2017learning,kool2018attention}, in which the graph is fully apparent and and the graph's edges are undirected. Conversely, multi-robot scheduling is a fundamentally different problem in which the graphical structure is a directed, acyclic graph with latent, disjointed temporal and spatial constraints that must be inferred.

\begin{figure*}[ht]
\centering
\includegraphics[width = 0.625\linewidth]{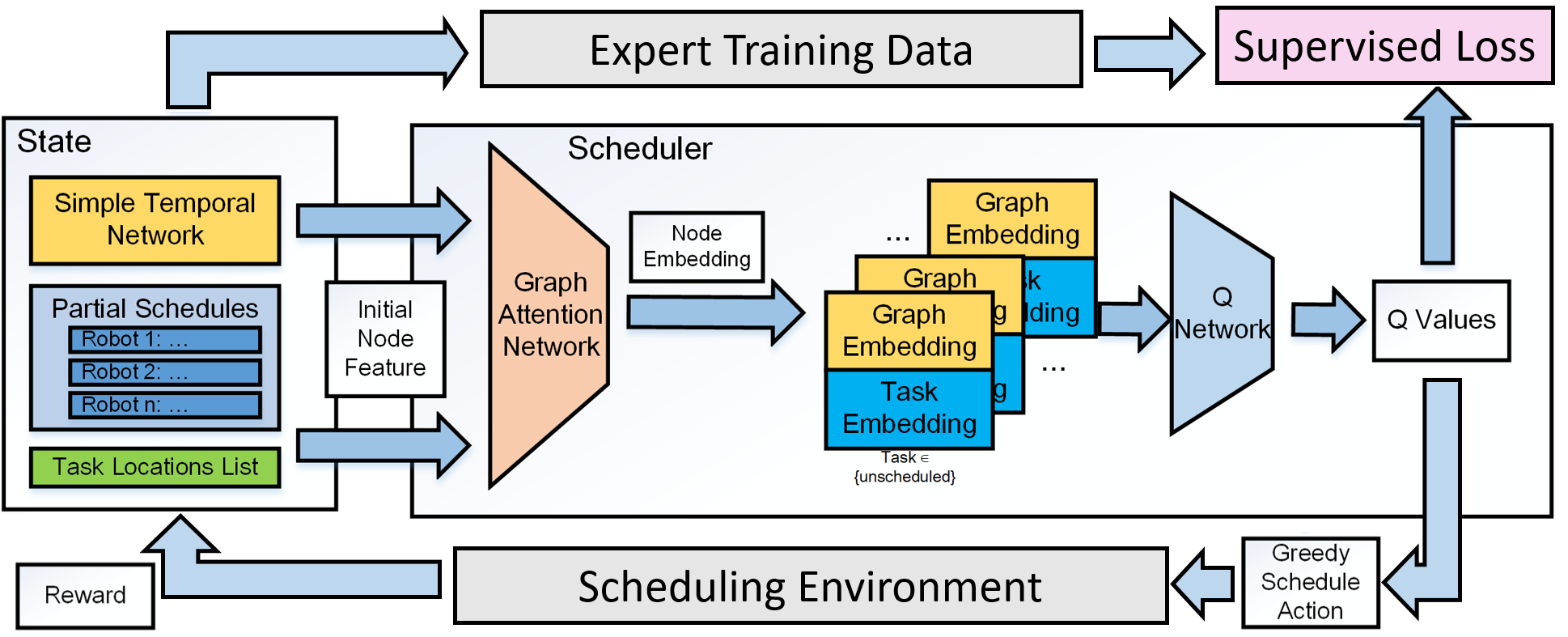}
\caption{The figure depicts the proposed framework, which incorporates graph attention networks and imitation and Q-learning for multi-robot scheduling.}
\label{fig:label1}
\end{figure*}

In this paper, we develop a novel graph attention network (GAT)~\cite{velickovic2018graph} model to learn scheduling policies that reason about the underlying simple temporal network (STN) structure~\cite{dechter1991temporal} and auxiliary constraints for multi-robot allocation and sequencing. Our GAT method is non-parametric in the number of tasks, meaning that the model can learn a policy from problem formulations of one size while still being able to construct schedules for task sets much larger than seen during training. This non-parametricity is relatively unique in machine learning but is fundamental to scheduling problems as the needs of the manufacturer evolve minute by minute. An added benefit is that our approach can leverage imitation learning from smaller problems in which supervised examples can be generated with exact solution methods, without the need for application-specific warm-start specifics, and still be fine-tuned with reinforcement learning on large-scale problems that are computationally intractable for exact approaches. We combine imitation learning with deep Q-learning to learn a heuristic policy for scheduling, allowing for fast, near-optimal scheduling of robot teams. The combined framework is illustrated in Fig.~~\ref{fig:label1}. We demonstrate that our approach is able to find solutions for \textgreater 90\% of problems for two robots scheduling up to 50 tasks with proximity constraints versus only 20\% by prior state-of-the-art methods.


\section{RELATED WORK}

Task assignment and scheduling for multi-robot teams is an important class of problems with applications to manufacturing, warehouse automation, and pickup-and-delivery~\cite{nunes2017taxonomy}. Our focus is on the single-task robots (ST), single-robot tasks (SR), time-extended assignment (TA) category with cross-schedule dependencies [XD] under the iTax taxonomy proposed by~\cite{korsah2013comprehensive}. Formulating the problem into a mixed-integer linear program (MILP) yields MILP-based solution techniques with exponential complexity, leading to intractability for factory operations\cite{brucker1999resource}. One popular way to accelerate the computation is to combine MILP and constraint programming (CP) methods into a hybrid algorithm using decomposition~\cite{benders1962partitioning, ren2009improved, jain2001algorithms}, but the performance may be limited by the decomposition quality. These approaches do not scale beyond a few agents and dozens of tasks.

Other hybrid approaches integrate heuristic schedulers within the MILP solver to achieve better scalability characteristics.~\cite{chen2009project} incorporated depth-first search (DFS) with heuristic scheduling.
Additional approaches perform cooperative scheduling by incorporating Tabu search within an MILP solver~\cite{tan2004linearized} or by applying heuristics to abstract the problem to groupings of agents~\cite{kushleyev2013towards}. Researchers have also sought to apply metaheuristic techniques, including simulated annealing (SA)~\cite{mousavi2013hybrid} and genetic algorithms (GAs)~\cite{zhang2015object} to specific scheduling problems.~\cite{dai2013energy} combined GAs and SA to further improve upon solution quality.

Some have pursued heuristic-learning for solving scheduling problems with approaches using policy~\cite{zhang1995reinforcement} and Q-learning~\cite{wang2005application,wu2011novel}. 
Yet, the common limiting factor of these methods is that they are either not multi-agent or they do not handle the robust set of temporal and spatial constraints that we consider (i.e., cross-schedule dependences [XD]). Moreover, these methods depend on customized features to achieve satisfying results.

To address these limitations, we consider recent advances in graph neural networks (GNN) that extend deep neural networks to handle arbitrarily-structured data~\cite{zhou2018graph}. 
Recently, GNNs have been used to solve combinatorial optimization problems, including the traveling salesman problem (TSP)~\cite{khalil2017learning, kool2018attention}. In these prior works, the node embeddings obtained from GNNs are combined with reinforcement learning algorithms to construct solutions. 
However, their models use undirected, unweighted graphs, and thus, are not suitable for multi-robot task allocation and scheduling problems, which use inherently directed, acyclic graphs often modeled as simple temporal networks (STNs)~\cite{barbulescu2010distributed, coltin2013online, nunes2015multi, gombolay2018fast}. 
To the best of our knowledge, we are the first to leverage GNNs in solving STN-based scheduling problems.


\section{PROBLEM STATEMENT}

We consider the problem of coordinating a multi-robot team in the same space, both with and without resource/location constraints. We describe its components, under the XD (ST-SR-TA) category of the widely accepted taxonomy proposed in~\cite{korsah2013comprehensive}, as a six-tuple $<r, \tau, d, w, Loc, z>$. $r$ are the robot agents that we assume are homogeneous in task completion. $\tau$ are the tasks to be performed. Each task $\tau_i$ is associated with a start time $s_i$ and a finish time $f_i$ and takes a certain amount of time $dur_i$ for each robot to complete. We introduce $s_0$ as the time origin and $f_0$ as the time point when all tasks are completed, so that the schedule has a common start and end point. $d$ is the deadline constraint that specifies a task has been completed before a certain time point. $w$ is the wait constraint that specifies the relative relationship between two tasks (e.g., ``task $i$ should wait at least 25 minutes after task $j$ finishes'' means $s_i \geq f_j + 25$). $Loc$ is the list of all task locations. At most, one task can be performed at each location at the same time. Finally, $z$ is an objective function to minimize that includes the makespan and possibly other application-specific terms.

A solution to the problem consists of an assignment of tasks to agents and a schedule for each agent's tasks such that all constraints are satisfied, and the objective function is minimized. We also include the mathematical program formation of our problem in Eq.~ \ref{eq:MILP_1}-\ref{eq:MILP_9}. We consider a generic objective function (Eq.~\ref{eq:MILP_1}), as application-specific goals vary. In Section \ref{sec:results}, we consider minimizing the makespan (i.e., overall process duration), which would be $z=\max_i f_i$.

Here we introduce two types of binary decision variables: 1) $A_{r,i}$ = 1 for the assignment of robot $r$ to task $i$ and 2) $X_{i,j}$ = 1 denotes task $i$ finishes before task $j$ starts. $L_{same}$ is the set of task pairs $(i, j)$ that use the same location and is derived from $Loc$. Eq.~\ref{eq:MILP_2} ensures that each task is assigned to only one agent. Eq.~ \ref{eq:MILP_3}-\ref{eq:MILP_5} ensure that all the temporal constraints are met. Eq.~ \ref{eq:MILP_6}-\ref{eq:MILP_7} ensure that robots can only perform one task at a time. Finally, Eq.~ \ref{eq:MILP_8}-\ref{eq:MILP_9} account for task locations that can only be occupied by one robot at a time. In Section \ref{sec:results}, we employ an exact benchmark (i.e., a mathematical program solver) to solve a linearized, mixed-integer form of these equations to improve computation time.\par\nobreak{\parskip0pt \normalsize \noindent
\begin{gather}
\min(z)\label{eq:MILP_1} \\
\sum_{r \in R}A_{r,i} = 1, \forall i \in \tau \label{eq:MILP_2}\\
f_i - s_i = dur_i, \forall i \in \tau\label{eq:MILP_3}\\
f_i - s_0 \leq d_i, \forall d_i \in \{d\}\label{eq:MILP_4}\\
s_i - f_j \geq w_{i,j}, \forall w_{i,j} \in \{w\}\label{eq:MILP_5}\\
(s_j - f_i)A_{r,i}A_{r,j}X_{i,j} \geq 0, \forall i,j \in \tau, \forall r \in R\label{eq:MILP_6}\\
(s_i - f_j)A_{r,i}A_{r,j}(1-X_{i,j}) \geq  0, \forall i,j \in \tau, \forall r \in R \label{eq:MILP_7}\\
(s_j - f_i)X_{i,j} \geq 0, \forall (i,j) \in L_{same}\label{eq:MILP_8}\\
(s_i - f_j)(1-X_{i,j}) \geq 0, \forall (i,j) \in L_{same} \label{eq:MILP_9}
\end{gather}}\noindent 


\section{REPRESENTATION: GRAPH NETWORKS}

Multi-robot task allocation and scheduling problems have been commonly modeled as STNs, because the consistency of the upper and lower bound temporal constraints can be efficiently verified in polynomial time. However, as we develop multiple agents, physical constraints, etc., we also have latent disjunctive variables that augment the graph to account for each agent being able to perform only one task at a time and for only one robot to occupy a work location at a time. This scheduling scenario is known as the Disjunctive Temporal Problem ~\cite{tsamardinos2003efficient}. GNNs are an ideal choice for reasoning about STNs given their graphical nature. However we must expand on prior work to handle both the directed nature of these graphs, as well as the disjunctive component from multi-robot coordination in time and space. These extensions are a key contribution of this paper.

Modern GNNs capture the dependence of graphs via message-passing between the nodes, in which each node aggregates feature vectors of its neighbors from previous layers to compute its new feature vector. After $k$ layers of aggregation, a node $v$'s representation captures the structural information within the nodes that are reachable from $v$ in $k$ hops or fewer. Systems based on GNNs have demonstrated ground-breaking performance on tasks such as node classification, link prediction, and clustering~\cite{zhou2018graph}. Here, we make use of the graph attention layer (GAT) proposed in~\cite{velickovic2018graph}, which is a variant of a graph convolutional layer that introduces an attention mechanism to improve generalizability and modify its structure to make it suitable for representing a STN.

\begin{figure*}[ht]
\centering
\subfigure[]{
    \begin{minipage}[t]{0.275\textwidth}
    \centering
    \includegraphics[width = \linewidth]{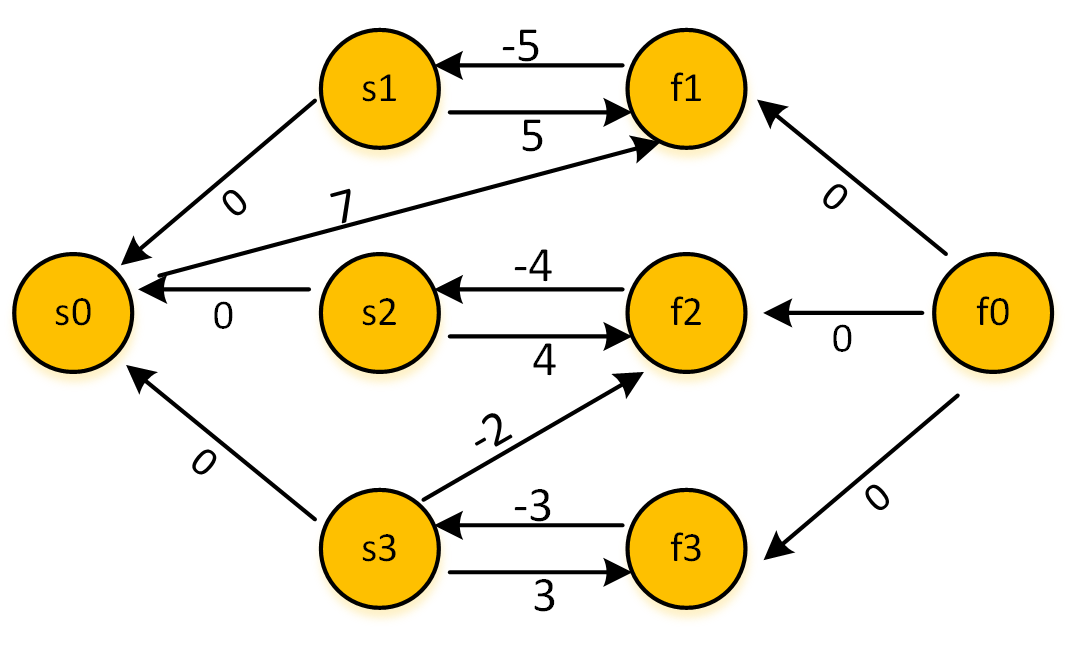}
    \label{fig:STN}
    \end{minipage}
}
\subfigure[]{
    \begin{minipage}[t]{0.575\textwidth}
    \centering
    \includegraphics[width = \linewidth]{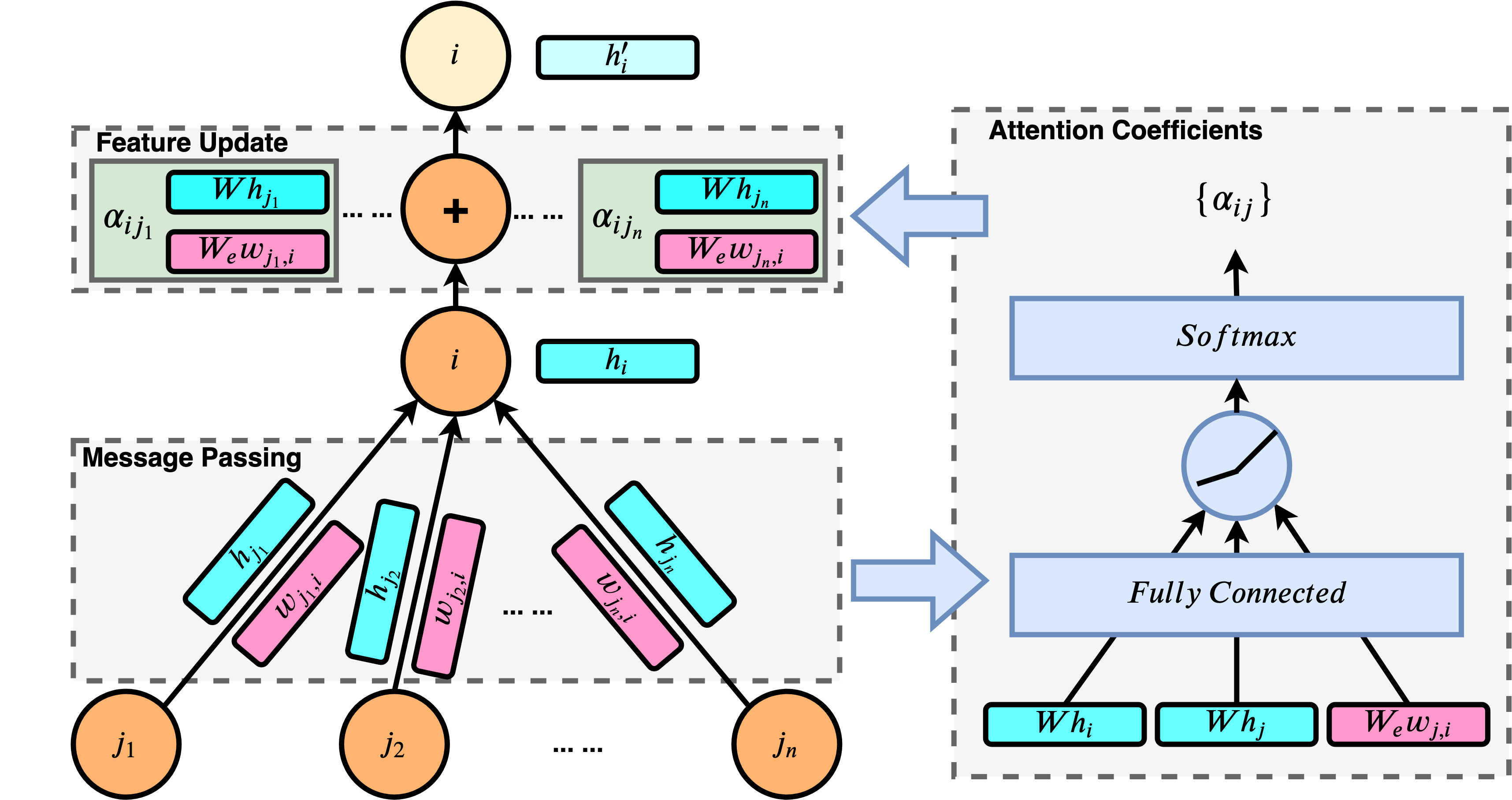}
    \label{fig:gatv2}
    \end{minipage}
}
\caption{Fig.~\ref{fig:STN} depicts an STN with start and finish nodes for three tasks, as well as placeholder start and finish nodes, $s_o$ and $f_o$. Task 1 has a deadline constraint and there is a wait constraint between task 2 and task 3. Fig.~\ref{fig:gatv2} depicts the forward pass of the adapted graph attention layer (left-hand side), which consists of two phases: 1) Message passing: each node receives features of its neighbor nodes and the corresponding edge weights; 2) Feature update: neighbor features are aggregated using attention coefficients; the right-hand side of Fig.~\ref{fig:gatv2} illustrates how attention coefficients are calculated.}
\end{figure*}

\textbf{Initial Node Features}--In a STN, each task $\tau_i$ is represented by two event nodes: its start time node $s_i$ and finish time node $f_i$. An example of a STN consisting of 3 tasks is shown in Fig.~\ref{fig:STN}. Given the partial schedule at the current step, we generate the initial input features of each node. The first N+1 dimensions are the one-hot encoding denoting the robot to which this task has been assigned, where N is the number of robots. For example, [0 1 0 ... 0] indicates the task is assigned to robot \# 2, and [0 ... 0 1] means that the task has not been assigned to any robot. We use [1 1 ... 1 0] to denote the placeholder start and finish nodes of the entire schedule, $s_o$ and $f_o$, respectively. The next two dimensions demonstrate whether this node represents the event start time [1 0] or finish time [0 1]. The next M dimensions are a one hot encoding of the location number the task uses, in which M is the number of total locations. In total, the input feature describing each node is an (M+N+3)-dimension binary vector, based on the current partial schedules of all robots. This set of input features is more expressive than that of prior approaches addressing the simpler TSP~\cite{khalil2017learning, kool2018attention} that only considered the (x,y) position of each node.

\newcommand{\norm}[1]{\left\lVert#1\right\rVert}
\textbf{Structure Adaptation:} The original graph attention network developed by~\cite{velickovic2018graph} is only able to incorporate undirected, unweighted graphs, yielding that model insufficient for scheduling problems in which temporal constraints are represented by the direction and weight of the edge between the two corresponding event nodes. As such, we make two improvements for the message passing and feature update phases as shown in Fig.~\ref{fig:gatv2}: 1) The message passing follows the same direction of the edge (i.e., only the incoming neighbors of a node are considered); 2) Edge information is also aggregated when updating the node feature, which is done by adding a fully-connected layer inside each GAT layer that transforms the edge weight $w$ into the same dimension as the node feature using $W_e$. As a result, the output node feature $\vec h_i^\prime$ is updated  by Eq.~ \ref{eq:GAT_layer}, where $N(i)$ is the set of neighbors of node i, $W$ is the weight matrix applied to every node, $\vec h_j$ is the node feature from the previous layer, and $\alpha_{ij}$ are the attention coefficients. To stabilize the learning process, we also utilize the multi-head attention extension, as described in~\cite{velickovic2018graph}. A multi-headed layer consists of $K$ independent GAT layers computing nodes features in parallel and concatenating those features as the output. \par\nobreak{\parskip0pt \normalsize \noindent \begin{equation}
\vec h_i^\prime = \mathrm{ReLU} \Big (\sum_{j \in N(i)} \alpha_{ij} (W \vec h_j+W_e w_{ji}) \Big )\label{eq:GAT_layer}
\end{equation}}\indent\textbf{Attention Coefficients}--The GAT layer computes the feature embedding for each node by weighting neighbor features from the previous layer with feature-dependent and structure-free normalization, which makes the network non-parametric in the number of tasks. The pair-wise normalized attention coefficients are computed as shown in Fig.~\ref{fig:gatv2} using Eq.~\ref{eq:attention}, where $\vec a$ is the attention weight to be learned, and edge features are also being utilized. LeakyReLU nonlinearity (with negative input slope $\alpha$ = 0.2) is used and followed by a softmax normalization.\par\nobreak{\parskip0pt \noindent \begin{equation}
\alpha_{ij} \sim e^{\mathrm{LeakyReLU}\left(\vec a^T\left[W \vec h_i\norm{W \vec h_j} W_e w_{ji}\right]\right)}\label{eq:attention}
\end{equation}}


\section{LEARNING SCHEDULING POLICY}

We first formulate scheduling as a sequential decision-making problem, in which individual robots' schedule are collectively, sequentially constructed from empty in a rollout fashion. At each decision step, the policy picks a robot-unscheduled task pair and assigns that unscheduled task to the end of that robot's schedule. This step repeats until all tasks are scheduled. We further impose the requirement that a robot which is assigned a task at a later step in the process cannot begin that task until all previously scheduled tasks have been started by their respective robots. Next, we formalize the problem of constructing the schedule as a Markov decision process (MDP) using a four-tuple $<x_t, u, T, R>$ that includes:

\begin{itemize}
\item{States: State $x_t$ at step $t$ indicates the graph updated with partial schedules of robots. As we encode each robot's partial schedule into the input feature of GNN, $x$ can be expressed as the graph embedding $h_g$ of all the partial schedules and is obtained by averaging over all node features extracted by the GNN.}
\item{Actions: Action $u$ = $<\tau_i$, $r_j>$ corresponds to appending an unscheduled task $\tau_i$ into the partial schedule of robot $r_j$. Task embedding $h_{\tau_i}$ is the average of the extracted features of start and end nodes, $s_i$ and $f_i$.}
\item{Transitions $T$: Transitions correspond to adding the edges associated with the action into the STN and updating the partial schedule of the selected robot.}
\item{Rewards $R$: The immediate reward of a state-action pair is defined as the change in makespan of all the scheduled tasks after taking the action. We divide the change by a discount factor $D > 1 $ if the next state is not a termination state. The reward is multiplied by -1.0, as we are minimizing the total makespan. A large negative reward $M_{inf}$ is returned if the action results in an infeasible schedule in the next state.}
\end{itemize}

We aim to learn a policy that schedules tasks and agents following the decision-making process. To enable combining imitation learning with deep Q-learning, we define an evaluation function, $Q(x_t, u_t)$, that calculates the total reward of taking action $u_t$ at step $t$. Then, our goal is to approximate the evaluation function with a neural network $\hat{Q}_\theta$ parameterized by weights $\theta$. $\hat{Q}_\theta$ takes as input the concatenation of graph embedding $h_g$ and task embedding $h_\tau$ and outputs $N$ scores estimating the total rewards of appending task $\tau$ to each of the $N$ robots' partial schedule. As a result, we obtain a greedy policy $\pi := \mathrm{argmax}_u \hat{Q}_\theta(h_g, h_\tau)$ that selects a task $\tau_i$ and a robot $r_j$ at each step to maximize the Q value with corresponding action. Fig.~\ref{fig:label1} illustrates the overall training architecture.

\textbf{Opportunistic Variant}--We also introduce a variant of the policy, which uses time-based rollout to generate the schedule. Starting from $t=0$ (here $t$ refers to time points instead of decision steps), at each time step, the policy first collects all the available robots not working on a task into a set $\boldsymbol{r_{avail}} = \{r_j | r_j \text{ is available}\}$. Then, $\forall r_j \in \boldsymbol{r_{avail}}$, the policy tries to assign $\tau_i$ using $\tau := \mathrm{argmax}_\tau \hat{Q}_\theta(h_g, h_\tau)|_{r=r_j}$ where only Q values from $r_j$ are used. To make this variant more robust, the schedule action is discarded if the lower bound of the picked task's start time is greater than current time. We compare both the baseline and opportunistic variant of our algorithm in Section \ref{sec:results}.

\begin{figure*}[ht]
    \centering
    \subfigure[]{ 
    \begin{minipage}[t]{0.3\textwidth}
        \centering
        \includegraphics[width = \linewidth,height = 0.62\linewidth]{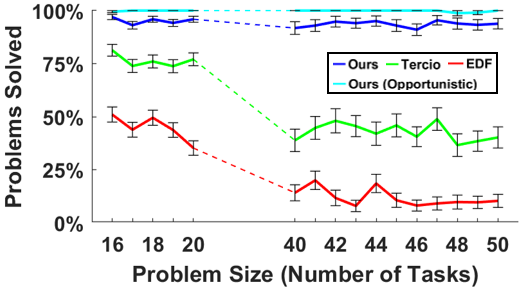}
        \label{fig:label4}
    \end{minipage}
    }
    \subfigure[]{
    \begin{minipage}[t]{0.3\textwidth}
    \centering
        \includegraphics[width = \linewidth,height = 0.62\linewidth]{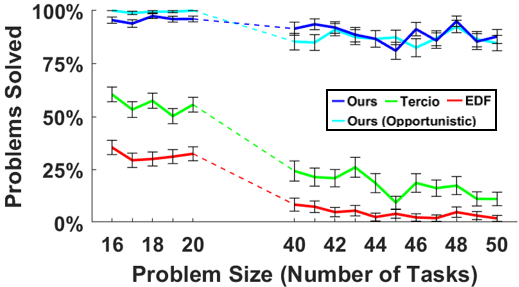}
        \label{fig:label5}
        \end{minipage}
    }
    \subfigure[]{
    \begin{minipage}[t]{0.3\textwidth}
    \centering
        \includegraphics[width = \linewidth,height = 0.62\linewidth]{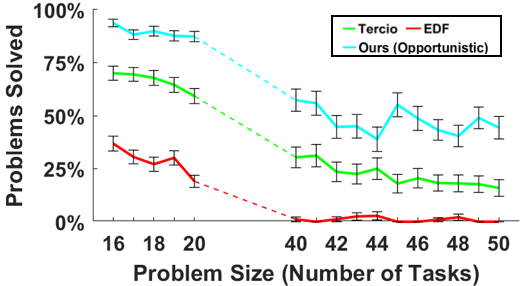}
        \label{fig:label6}
        \end{minipage}
    }
    \caption{Proportion of problems solved for two robots without (a) and with (b) location constraints, as well as for five robots with location constraints.}
    \label{fig:proportions}
\end{figure*}


\subsection{Imitation Learning}
Although obtaining optimal solutions of large-scale scheduling problems is computationally intractable, it is practical to optimally solve smaller-scale problems with exact methods. Furthermore, we can use these exact methods to automatically generate application-specific examples for training an imitation learning algorithm without the need for the tedious, non-trival task of developing application-specific heuristics to warm-start the solver. Finally, we typically have access to high-quality, manually-generated schedules from human experts that currently manage the logistics in manufacturing environments. We believe that exploiting such expert data to train the scheduling policy can greatly accelerate the learning process~\cite{piot2014boosted}.

We aim to leverage such data by initially training the network on expert dataset $D_{ex}$ that contains all the state-action pairs of schedules either from exact solution methods or the domain experts. For each transition, we directly calculate the total reward from current step $t$ until termination step $n$ using $R_t^{(n)} = \sum_{k=0}^{n-t} \gamma^k R_{t+k}$ and regress $\hat{Q}_\theta$ towards this value as shown in Eq.~\ref{eq:L_ex}, where the supervised learning loss, $L_{ex}$, is defined as the Euclidean distance between the $R_t^{(n)}$ and our current estimate based on graph embedding $h_g$ and embedding of the task selected by the expert $h_{\tau, ex}$. \par\nobreak{\parskip0pt \normalsize \noindent
\begin{equation}
L_{ex} = \norm{\hat{Q}_\theta(h_g,h_{\tau,ex})-R_t^{(n)}}^2 \label{eq:L_ex}
\end{equation}}\indent To fully exploit the expert data, we ground the Q values of actions that are not selected by the expert to a value below $R_t^{(n)}$ using the loss shown in Eq.~\ref{eq:_L_alt}, where $h_{\tau,alt}$ is the task embedding associated with alternate actions not chosen by the expert, $C$ is a positive constant used as an offset, and $N_{alt}$ is the number of alternate actions at step $t$.  \par\nobreak{\parskip0pt \normalsize \noindent \begin{align}
L_{alt} &= \frac{1}{N_{alt}}\sum \left\lVert\hat{Q}_\theta(h_g,h_{\tau,alt})\right. \nonumber\\
&\indent\indent \indent\indent \left.-\min(\hat{Q}_\theta(h_g,h_{\tau,alt}), R_t^{(n)} - q_o)\right\rVert^2 \label{eq:_L_alt}
\end{align}}Consequently, the gradient propagates through all the unselected actions that have Q values higher than $R_t^{(n)}-q_o$. We select $q_o$ empirically during training. Note the difference from~\cite{piot2014boosted} in that they only train on the unselected action with the max Q value. The total supervised loss is shown in Eq.~\ref{eq:L_sup}, where $L_2$ is the L2 regularization term on the network weights, and $\lambda_1, \lambda_2$ are weighting parameters assigned to different loss terms empirically. \par\nobreak{\parskip0pt \normalsize \noindent \begin{equation}
L_{sup} = L_{ex} + \lambda_1 L_{alt} + \lambda_2 L_2  \label{eq:L_sup}
\end{equation}}


\begin{figure*}[ht]
    \centering
    \subfigure[]{ 
    \begin{minipage}[t]{0.3\textwidth}
        \centering
        \includegraphics[width = \linewidth,height = 0.62\linewidth]{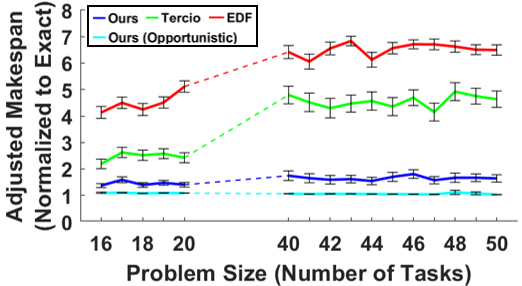}
        \label{fig:label7}
    \end{minipage}
    }
    \subfigure[]{
    \begin{minipage}[t]{0.3\textwidth}
    \centering
        \includegraphics[width = \linewidth,height = 0.62\linewidth]{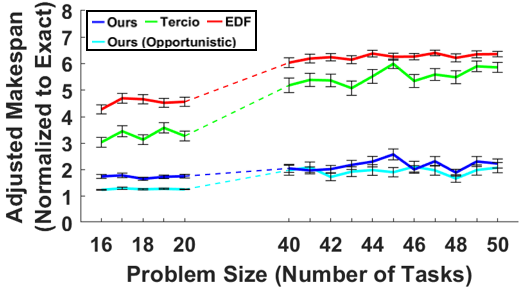}
        \label{fig:label8}
        \end{minipage}
    }
    \subfigure[]{
    \begin{minipage}[t]{0.3\textwidth}
    \centering
        \includegraphics[width = \linewidth,height = 0.62\linewidth]{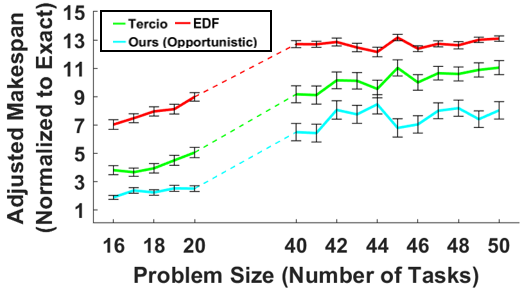}
        \label{fig:label9}
        \end{minipage}
    }
    \caption{Adjusted makespan score for two robots without (a) and with (b) location constraints, as well as for five robots with location constraints. A smaller adjusted makespan is better, with the optimal value normalized to one.}
    \label{fig:adjustedMakespans}
\end{figure*}

\subsection{Deep Q-learning}
This approach to imitation learning with expert data helps the policy to encapsulate the decision-making characteristics of exact algorithms on a smaller scale (or domain experts with high-quality, but suboptimal schedules on a larger scale) in a data-efficient manner. However, in practice, we find that the performance of a trained policy often decreases after a certain amount of training steps, not only on the testing data but also on the training data. We argue that having $q_o$ is beneficial in the short-run, but relying on $q_o$ ultimately introduces too much ``bias'' in a way that prohibits finding the desired, latent Q-function.

To address this issue, after initial pre-training solely via imitation learning, we transition to training the model with self-play data $D_{selfplay}$ generated by running an $\epsilon$-greedy version of the trained policy on the training problems. This training process falls into the classic deep Q-learning framework~\cite{hessel2018rainbow}. The loss for self-play data uses a 1-step reward with double Q-learning~\cite{van2016deep} and is calculated as shown in Eq.~\ref{eq:LDQN}, where $\hat{Q}_{\theta^\prime}^\prime$ is the target network, and $\tau^* = \mathrm{argmax}_{\tau^\prime}\hat{Q}_\theta(h_{g,t+1},h_{\tau^\prime})$.  \par\nobreak{\parskip0pt \normalsize \noindent \begin{equation}
L_{dqn} = \left\lVert R_t + \gamma \hat{Q}_{\theta^\prime}^\prime\left(h_{g,t+1}, h_{\tau^*}\right) - \hat{Q}_\theta(h_{g,t},h_{\tau})\right\rVert^2
\label{eq:LDQN}
\end{equation}}For small-scale problems, we combine $L_{dqn}$ with the supervised loss to further train the network, which yields better performance. In this case, we assign a weighting parameter $\lambda_3$ to $L_{dqn}$ before adding it into Eq.~\ref{eq:L_sup}. For larger-scale problems, where expert data is not available, we use $L_{dqn}$ to fine-tune the network after initializing with weights trained on smaller-scale problems.


\section{RESULTS AND DISCUSSION}
\label{sec:results}
We evaluate the performance of our model on randomly-generated problems simulating multi-agent construction of a large workpiece, e.g. an airplane fuselage. We generate problems involving a team of two robots in different scales: small (16--20 tasks) and large (40--50 tasks), both without and with proximity/location constraints (i.e., no two robots can be in the same location at the same time). We also generate problems involving a team of five robots with location constraints. Task duration is generated from a uniform distribution in the interval $[1, 10]$. In keeping with distributions typically found in manufacturing environments, approximately 25\% of the tasks have absolute deadlines drawn from a uniform distribution in the interval [1, $T$ * 5], where $T$ is the number of total tasks. Approximately 25\% of the tasks have wait constraints; the duration of non-zero wait constraints is drawn from a uniform distribution in the interval $[1, 10]$. We set the number of locations to be the same as the number of robots, and each task's location is picked randomly. For small and large problems of each robot setting, we generate 1,000 training and 1,000 testing problems. For supervised learning on small problems, Gurobi provides expert schedules, with a cutoff time of two minutes for two-robot problems and 15 minutes for five-robot problems.


\textbf{Model Details}--Our code implementation uses pyTorch~\cite{paszke2017automatic}, and the graph neural networks are built upon Deep Graph Library (https://www.dgl.ai). We apply a three-layer GAT to learn node features. Each layer uses 8 attention heads computing 64 features. The last GAT layer uses averaging while the first two use concatenation to aggregate the features from each head. The Q network uses two fully-connected layers with a hidden dimension of 64. We preprocess the STN using Floyd Warshall's algorithm to find its minimum distance graph as the input graph~\cite{dechter1991temporal}.
We set $\gamma$ = 0.95 and use Adam optimizer~\cite{kingma2014adam} through training. Imitation learning uses a learning rate of $10^{-5}$, $\lambda_1$ = 0.8, and $\lambda_2$ = 0.1. Deep Q-learning uses a learning rate of $10^{-6}$, $\lambda_1$ = 0.8, $\lambda_2$ = 0.1, $\lambda_3$ = 2, and $\epsilon$ = 0.05. The target Q network is updated every 5,000 steps.

\textbf{Benchmarks}--We benchmark our model against a ubiquitous heuristic algorithm, earliest deadline first (EDF), as well as the state-of-the-art scheduling algorithm for this problem domain, Tercio~\cite{gombolay2018fast}. EDF works by selecting from a list of available tasks the one with the earliest deadline and assigns it to the first available worker. Tercio is a hybrid algorithm combining mathematical optimization for task allocation and analytical sequencing test to ensure temporal and spatial feasibility. Results are normalized to the exact solution.

We evaluate our model on two metrics: 1) Proportion of problems solved and 2) Adjusted makespan. To calculate the adjusted makespan, we assign a value, $20T$, as the makespan of problems not solved by the algorithm, which was the reward signal we employed to train our algorithm. We employ a model ensemble (size = 3) during testing, in which we run 3 models picked from different training steps on the same problem and use the schedule with the minimum makespan as the final output. Fig.~\ref{fig:proportions} shows the proportion of problems solved of our approach and for the baseline methods for problems with two robots without (Fig.~\ref{fig:label4}) and with (Fig.~\ref{fig:label5}) location constraints and for five robots with location constraints (Fig.~\ref{fig:label6}). Fig.~\ref{fig:adjustedMakespans} shows the adjusted makespan score, normalized to the value found by the exact method, of our approach and for the baseline methods for problems with two robots without (Fig.~\ref{fig:label7}) and with (Fig.~\ref{fig:label8}) location constraints and for five robots with location constraints (Fig.~\ref{fig:label9}).

For scheduling two robots without location constraints, the opportunisitic variant achieved the best results (99.8\% of problems solved with 16--20 tasks and 99.8\% for problems with 40--50 tasks). While this variant also achieved the best results for small problems involving two robots with locations constraints, the decision step-based counterpart achieved slightly higher performance on large problems for two robots with location constraints (86.4\% vs. 84.5\%). However, the opportunistic variant still performed the best in terms of adjusted makespan. Our trained policy showed  consistent, high-performance among different problem sizes, while the performance of EDF and Tercio decreased precipitously when the number of tasks increased (e.g., in Fig.~\ref{fig:label5}, proportion of problems solved dropped from 54.1\% on small problems to 17.2\% on large problems for Tercio). It should be noted that we only used expert data on small problems during training and relied on Q-learning for fine-tuning on the larger problem sizes. This positive result provides promising evidence that our framework is able to transfer knowledge learned on small problems to help solve large problems. We further evaluated the capability of the opportunistic variant on more difficult problems with five robots and locations constraints in which our algorithm was able to outperform Tercio and EDF (Fig.~\ref{fig:label6} and \ref{fig:label9}).

Finally, to show the necessity and benefit of incorporating edge information into the GAT layer, we also trained and evaluated a policy based on the original GAT models, using small problems containing two robots. As expected, the trained policy performed only solved 12.8\% and 5.7\% of the problems without and with location constraints, respectively. These results further show the power of our approach leveraging graph attention networks to automatically learn to coordinate robot teams in complex scheduling environments.


\section{CONCLUSIONS}

We presented a graph-attention-network-based framework that can automatically learn a scalable scheduling policy to coordinate multi-robot teams. By combining imitation learning on small problems with deep Q-learning on larger problems in a non-parametric framework, we were able to obtain a deterministic, greedy policy that generated fast, near-optimal scheduling of robot teams. We demonstrated that our network-based policy found at least twice as many solutions over prior state-of-the-art methods in all testing scenarios.



\addtolength{\textheight}{-6.5cm}   







\bibliographystyle{IEEEtran}
\bibliography{IEEEabrv,example.bib}

\end{document}